\documentclass[runningheads]{llncs}
\usepackage{graphicx}
\usepackage{amsmath}
\usepackage{amsfonts} 
\usepackage{graphicx}
\usepackage{graphbox}
\usepackage{booktabs}       % professional-quality tables
\begin{document}
\title{Recurrent Neural Networks for Forecasting Time Series with Multiple Seasonality: A Comparative Study}

%Recurrent Neural Networks for Pattern-based Short-Term Load Forecasting: A Comparative Study}

\titlerunning{RNN for Forecasting TS with Multiple Seasonality: A Comparative Study}

%RNN for Pattern-based STLF: A Comparative Study}
\author{Grzegorz Dudek \inst{1}\orcidID{0000-0002-2285-0327}
\and
Slawek Smyl\inst{2}\orcidID{0000-0003-2548-6695} \and
Paweł Pełka\inst{1}\orcidID{0000-0002-2609-811X}}
\authorrunning{G. Dudek et al.}
\institute{Electrical Engineering Faculty, Czestochowa University of Technology, Poland\\
\email{\{grzegorz.dudek,pawel.pelka\}@pcz.pl} \and
%Meta, 1 Hacker Way, Menlo Park, CA 94025, USA\\
\email{slawek.smyl@gmail.com} 
}
\maketitle              % typeset the header of the contribution
\begin{abstract}
This paper compares recurrent neural networks (RNNs) with different types of gated cells for forecasting time series with multiple seasonality. The cells we compare include classical long short term memory (LSTM), gated recurrent unit (GRU), modified LSTM with dilation, and two new cells we proposed recently, which are equipped with dilation and attention mechanisms. To model the temporal dependencies of different scales, our RNN architecture has multiple dilated recurrent layers stacked with hierarchical dilations. The proposed RNN produces both point forecasts and predictive intervals (PIs) for them. An empirical study concerning short-term electrical load forecasting for 35 European countries confirmed that the new gated cells with dilation and attention performed best.

\keywords{LSTM \and Multiple seasonality \and RNN \and Short-term load forecasting \and Time series forecasting.}
\end{abstract}

\section{Introduction}

Forecasting time series (TS) with multiple seasonality is a challenging problem. To solve it, a forecasting model has to deal with short- and long-term dynamics as well as a trend and variable variance. Classical statistical methods such as autoregressive moving average (ARMA) and exponential smoothing methods can be extended to multiple seasonal cycles \cite{Box94,Tay10} but they suffer from many drawbacks. The most important of these are: their linear nature, limited adaptability, limited ability to model complex seasonal patterns, problems with capturing long-term dependencies and problems with introducing exogenous variables.  

To improve the ability of statistical models to capture multiple seasonality, various approaches have been applied, such as extending the model with Fourier terms \cite{Liv11,Tay18}, TS decomposition \cite{Hov15} and local modeling \cite{Dud16,Sha20}. Machine learning (ML) gives additional opportunities to the models and makes them more flexible. The main idea behind ML is to learn from past observations any inherent structures, patterns or anomalies within the data, with the objective of generating future values for the series. The most popular ML models in the field of forecasting are neural networks (NNs) \cite{Ben20} as they can flexibly model complex nonlinear relationships
with minimum a-priori assumptions and reflect process variability in uncertain dynamic environments. They offer learning of representation, cross-learning on massive datasets and modeling temporary relationships in sequential data. In particular, RNNs, which were designed for sequential data such as TS and text data, are extremely useful for forecasting. They form a directed graph along a temporal sequence which is able to exhibit temporal dynamic behavior using their internal state (memory) to process sequences of inputs. 

Modern RNNs, such as LSTM and GRU, are capable of learning both short and long-term dependencies in TS \cite{Hew21}. They are equipped with recurrent cells that can maintain their states over time and, using nonlinear “regulators” called gates, can control the flow of information inside the cell. Recent works have reported that gated RNNs provide high accuracy in forecasting and outperform most of the statistical and ML methods, such as
ARIMA, support vector machine, and shallow NNs \cite{Yan18}. A comparison of RNNs on multiple seasonality forecasting problems performed in \cite{Bia17} showed that LSTM, GRU and classical Elman RNN demonstrate comparable performance but are relatively slow in terms of training time due to the time-consuming backpropagation through the time procedure. To improve the learning capability and forecasting performance facing RNN, different mechanisms have been used such as  residual connections \cite{He16} and dilated architecture \cite{Cha17}, which solves the major challenges of RNN when learning on long sequences: i.e. complex dependencies, vanishing and exploding gradients, and efficient parallelization. Hybrid solutions have also been proposed combining RNN with TS decomposition \cite{Ban22} or other methods such as exponential smoothing. One such hybrid model won the reputed M4 forecasting competition in 2018, showing impressive performance \cite{Smy20}.

%"Forecasting with Multiple Seasonality"
%"Forecasting Time Series with Multiple Seasonal Patterns"
%Recurrent Neural Networks for Time Series Forecasting G´abor Petneh´azi∗
%Ensembles of Recurrent Neural Networks for Robust Time Series Forecasting Sascha Krstanovic and Heiko Paulheim
%Forecasting with Multiple Seasonality, https://arxiv.org/pdf/2008.12340.pdf
%Mixed pooling of seasonality for time series forecasting: An application to pallet transport data

Motivated by the superior performance of RNN in TS forecasting, in this study, we compare RNNs with different recurrent cells. We consider a problem of univariate forecasting TS with multiple seasonality on the example of short-term electrical load forecasting (STLF). We propose a stacked hierarchical RNN architecture trained globally across all series and equipped with recurrent cells of different types. 
We normalize TS input data and encode output data using coding variables determined from recent history. This is to better capture the current dynamics of the process. Such prepossessing has proven successful in other forecasting models for multiple seasonality, see \cite{Dud16,Dud16a,Dud20}. 

The contribution of this study is as follows:
\begin{enumerate}
    \item 
    We propose a new RNN architecture for forecasting TS with multiple seasonality. It is composed of three dilated recurrent layers stacked with hierarchical dilations to deal with multiple seasonality. It uses a combined asymmetrical loss function which enables the model to produce both point forecasts and PIs and also to reduce the forecast bias. 
    \item
    We compare five types of gated recurrent cells: classical LSTM and GRU, modified LSTM with dilation, and two new cells we proposed recently, which are equipped with dilation and attention mechanisms.
    \item 
    We empirically demonstrate on real data for the electricity demand for 35 European countries that our proposed model copes successfully with complex seasonality. The new attentive dilated recurrent cell significantly outperforms its competitors in terms of accuracy.
\end{enumerate} 

The remainder of the paper is organised as follows. Section 2 describes the forecasting problem and data representation. Section 3 presents the recurrent cells and Section 4 describes RNN architecture. Section 5 describes the results of experiments and discusses our findings. Finally, Section 6 concludes the paper.

\section{Forecasting Problem and Data Representation}

In this study, as an example of forecasting time series with multiple seasonality, we consider a problem of STLF. The hourly load time series, $\{z_\tau\}_{\tau=1}^M$, express triple seasonality: yearly, weekly and daily (see \cite{Smy21} for details, where such time series are analysed). Our goal is to forecast the daily profile (24 hours) for the next day based on historical loads (univariate problem).

As input information, we introduce a weekly profile, which precedes the forecasted day. This profile is represented by the input pattern defined as follows:

\begin{equation}\label{eqx}
\mathbf{x}_t = \frac{\mathbf{z}^w_t-\overline{z}^w_t}{\text{std}({z}^w_t)}
\end{equation}
where $\mathbf{x}_t \in \mathbb{R}^{168}$ is the $t$-th weekly pattern, $\mathbf{z}^w_t\in \mathbb{R}^{168}$ is the original sequence of the $t$-th week, and $\overline{z}^w_t$ and $\text{std}({z}^w_t)$ are its mean and standard deviation, respectively.

Note that \eqref{eqx} expresses standardization of the weekly sequence. Thus the weekly sequences for $t=1,..., N$ are unified, i.e. they are centered around zero with a unit variance. This operation filters out the trend and yearly seasonality.

An output pattern represents a forecasted daily sequence as follows:
\begin{equation}\label{eqy}
\mathbf{y}_t = \frac{\mathbf{z}^d_t-\overline{z}^w_t}{\text{std}({z}^w_t)}
\end{equation}
where $\mathbf{y}_t\in \mathbb{R}^{24}$ is the $t$-th daily pattern and $\mathbf{z}^d_t \in \mathbb{R}^{24}$ is the forecasted sequence (following directly weekly sequence $\mathbf{z}^w_t$ which is encoded in $\mathbf{x}_t$).

Note that in \eqref{eqy}, we encode the daily sequence using the mean and standard deviation of the preceding week. This enables us to decode the forecasted pattern, $\hat{\mathbf{y}}$, into the real sequence as follows:

\begin{equation}\label{eqz}
\hat{\mathbf{z}}^d = \hat{\mathbf{y}}\text{std}({z}^w) + \overline{z}^w
\end{equation}
where $\overline{z}^w$ and $\text{std}({z}^w)$ are coding variables determined on the basis of the historical weekly sequence represented by query pattern $\mathbf{x}$.

Following \cite{Smy21}, to introduce more input information related to the forecasted sequence, we extend the input vector with the following components: $\log_{10}(\bar{z}^w_t)$, which informs about the level of the time series, $\textbf{d}_t^{w} \in \{0, 1\}^7, \textbf{d}_t^{m} \in \{0, 1\}^{31}$ and $\textbf{d}_t^{y} \in \{0, 1\}^{52}$, which are binary one-hot vectors encoding day of the week, day of the month and week of the year for the forecasted day. The extended input pattern takes the form: 

\begin{equation}
\textbf{x}_t'= [\textbf{x}_t,\, \log_{10}(\bar{z}_t),\, \textbf{d}_t^{w},\, \textbf{d}_t^{m},\, \textbf{d}_t^{y}] 
\label{eqxp}
\end{equation} 

The paired input and output patterns constitute the training set, $\{(\mathbf{x}'_i,\mathbf{y}_i)\}^N_{i=1}$. The proposed model is trained in cross-learning mode, i.e. on many time series \cite{Smy20}, which enables it to capture the shared features of the individual series and prevents over-fitting. The training sets for all $L$ time series are combined: $\Phi=\Phi_1 \cup ... \cup \Phi_L$.  

\section{Recurrent Cells}
\label{RC}

In our study, we explore RNNs with different gated recurrent cells. They include classical cells such as LSTM and GRU, modified LSTM, i.e. dilated LSTM (dLSTM), and two new solutions proposed recently, dRNNCell and adRNNCell. 

\subsection{LSTM}

LSTM was proposed in \cite{Hoch97}  for learning problems related to sequential data. The main idea behind LSTM is a memory cell that carries relevant information throughout the processing of the sequence, and nonlinear gating units that regulate the information flow in the cell. Due to the memory, long-term temporal relationships can be captured and the effects of short-term memory can be reduced, i.e. even information from the earlier time steps can make its way to later time steps. Moreover, in LSTM, unlike in simple RNNs, the optimization problem with vanishing gradients was reduced, which improved learning capabilities. 

\begin{figure}[h]
    \begin{minipage}{1\textwidth}
        \centering
        \includegraphics[align=c,width=0.4\textwidth]{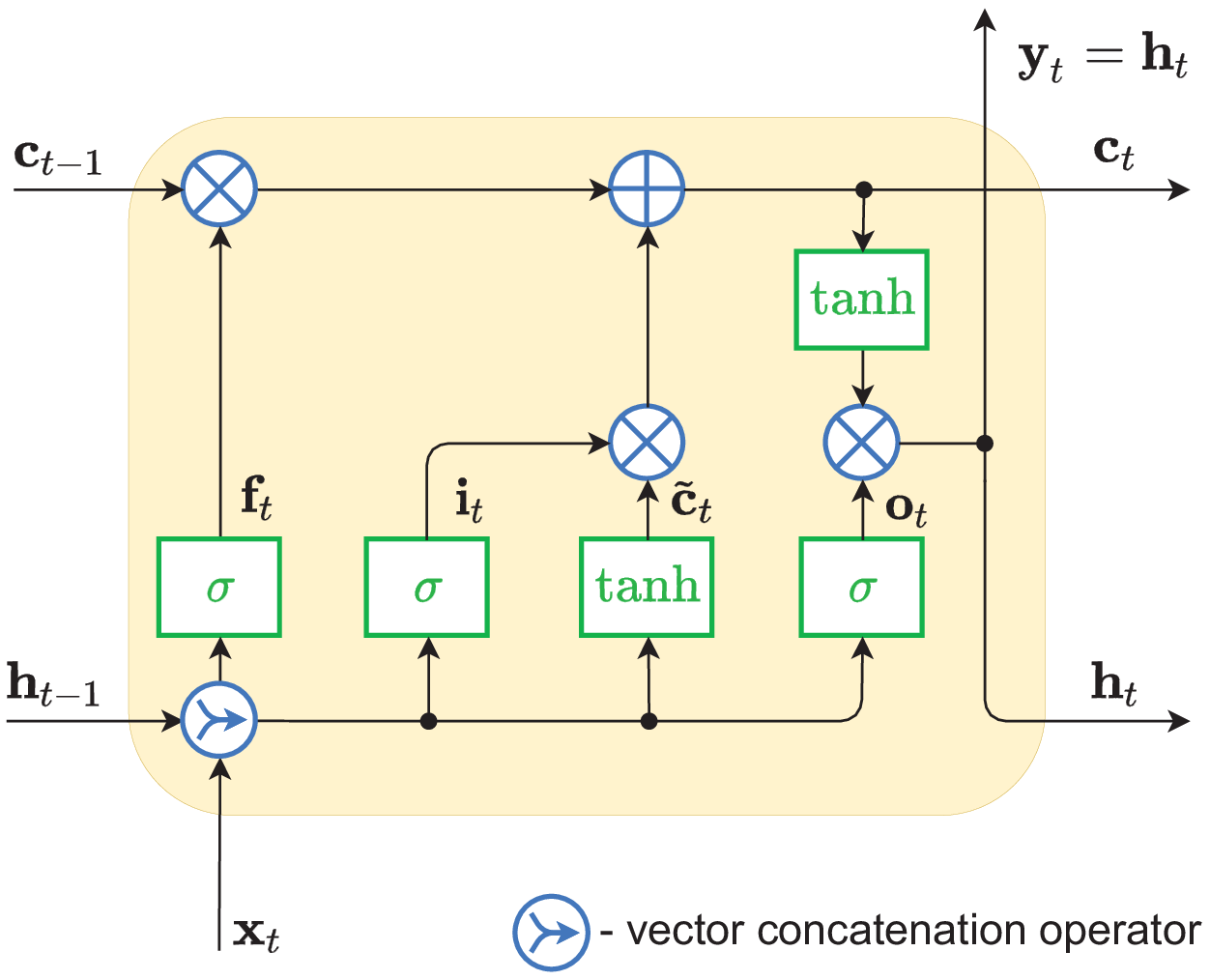}
        \hspace*{0.1cm}
        \includegraphics[align=c,width=0.4\textwidth]{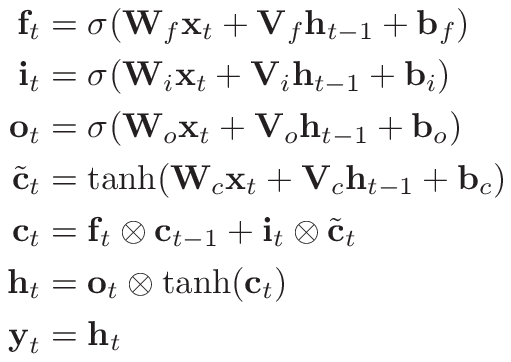}
    \end{minipage}
    \caption{LSTM.}
    \label{figLSTM}
\end{figure}

Fig. \ref{figLSTM} shows a diagram of LSTM. LSTM uses two states: a cell state,  $\textbf{c}_t$, and a hidden state,  $\textbf{h}_t$. The states contain information learned from the previous time steps. At each time step $t$, information is added to or removed from the cell state. These updates are controlled using three gates, which in fact are layers of learned nonlinear transformations. They comprise: input gate ($i$), forget gate ($f$) and output gate ($o$). All of the gates receive the hidden state of the past cycle and the current time series sequence as inputs.
They can learn what information is relevant to keep or forget during training.
At time step $t$, the cell uses the recent states, $\textbf{c}_{t-1}$ and $\textbf{h}_{t-1}$, and the input sequence, $\textbf{x}_{t}$, to compute new updated states $\textbf{c}_{t}$ and $\textbf{h}_{t}$.
The hidden and cell states are recurrently connected back to the cell input. The new hidden state, $\textbf{h}_{t}$, has two functions. It controls the gating mechanism in the next step and it is treated as the cell output, $\textbf{y}_{t}$, which goes to the next layer.  

The compact form of the equations describing LSTM are shown in Fig. \ref{figLSTM}, where:
$\textbf{W}$ and $\textbf{V}$ are learned weight matrices, $\textbf{b}$ are learned bias vectors, $\otimes$ denotes the Hadamard product and $\sigma$ is a logistic sigmoid function.

\subsection{GRU}

\begin{figure}[h]
    \begin{minipage}{1\textwidth}
        \centering
        \includegraphics[align=c,width=0.37\textwidth]{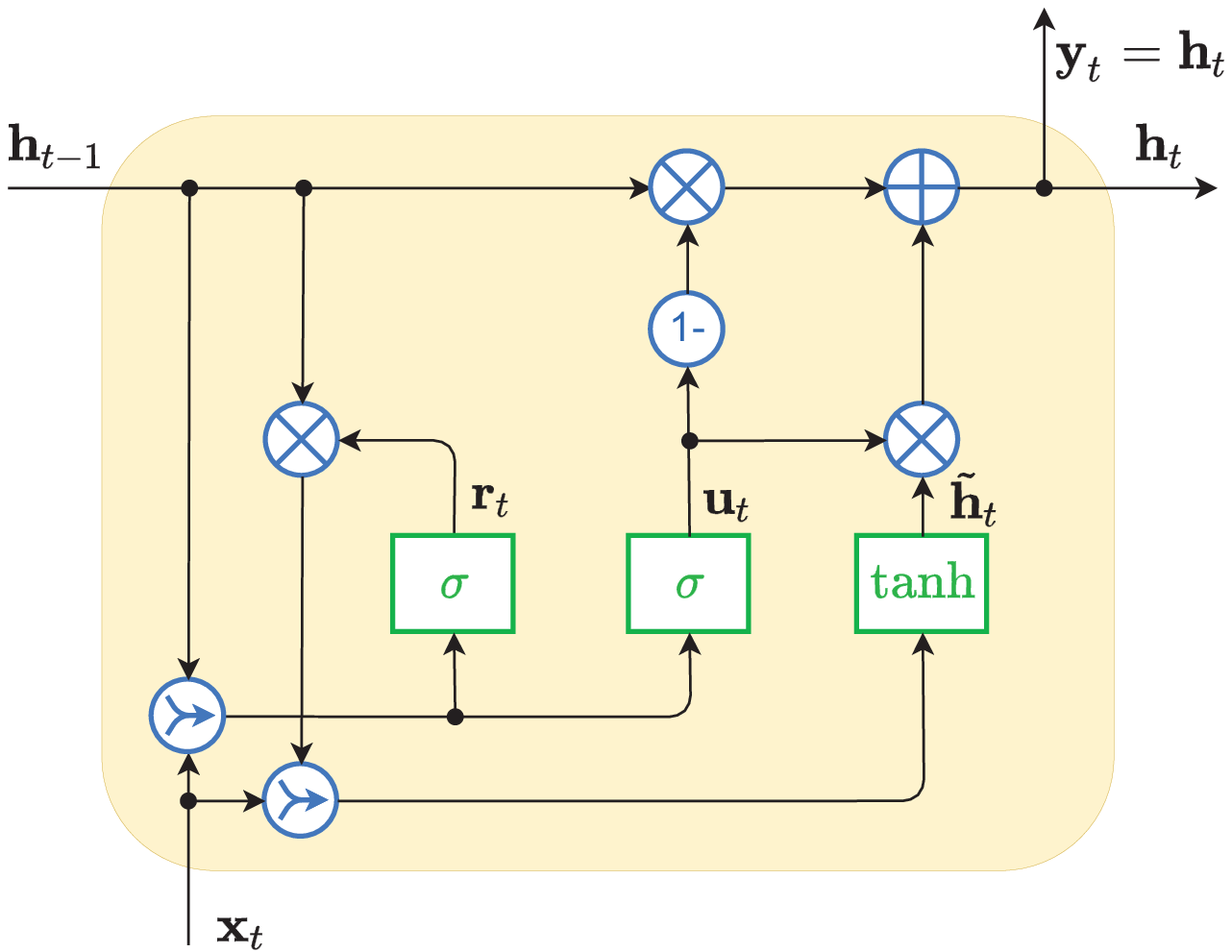}
        \hspace*{0.0cm}
        \includegraphics[align=c,width=0.50\textwidth]{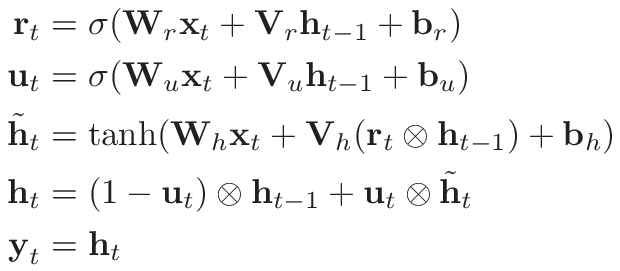}
    \end{minipage}
    \caption{GRU.} 
    \label{figGRU}
\end{figure}

In comparison to LSTM, in GRU the cell state was eliminated so the hidden state is used to both store information and control the gating mechanism \cite{Cho14}. 
GRU only has two gates, a reset gate ($r$) and an update gate ($u$). The update gate acts in a similar way as the forget and input gates in LSTM. It decides what information to remove and what new information to add. The reset gate decides how much past information to forget. The output gate was eliminated. The gating mechanism of GRU and the corresponding equations are shown in Fig. \ref{figGRU}. 

\subsection{dLSTM}

\begin{figure}[h]
    \begin{minipage}{1\textwidth}
        \centering
        \includegraphics[align=c,width=0.40\textwidth]{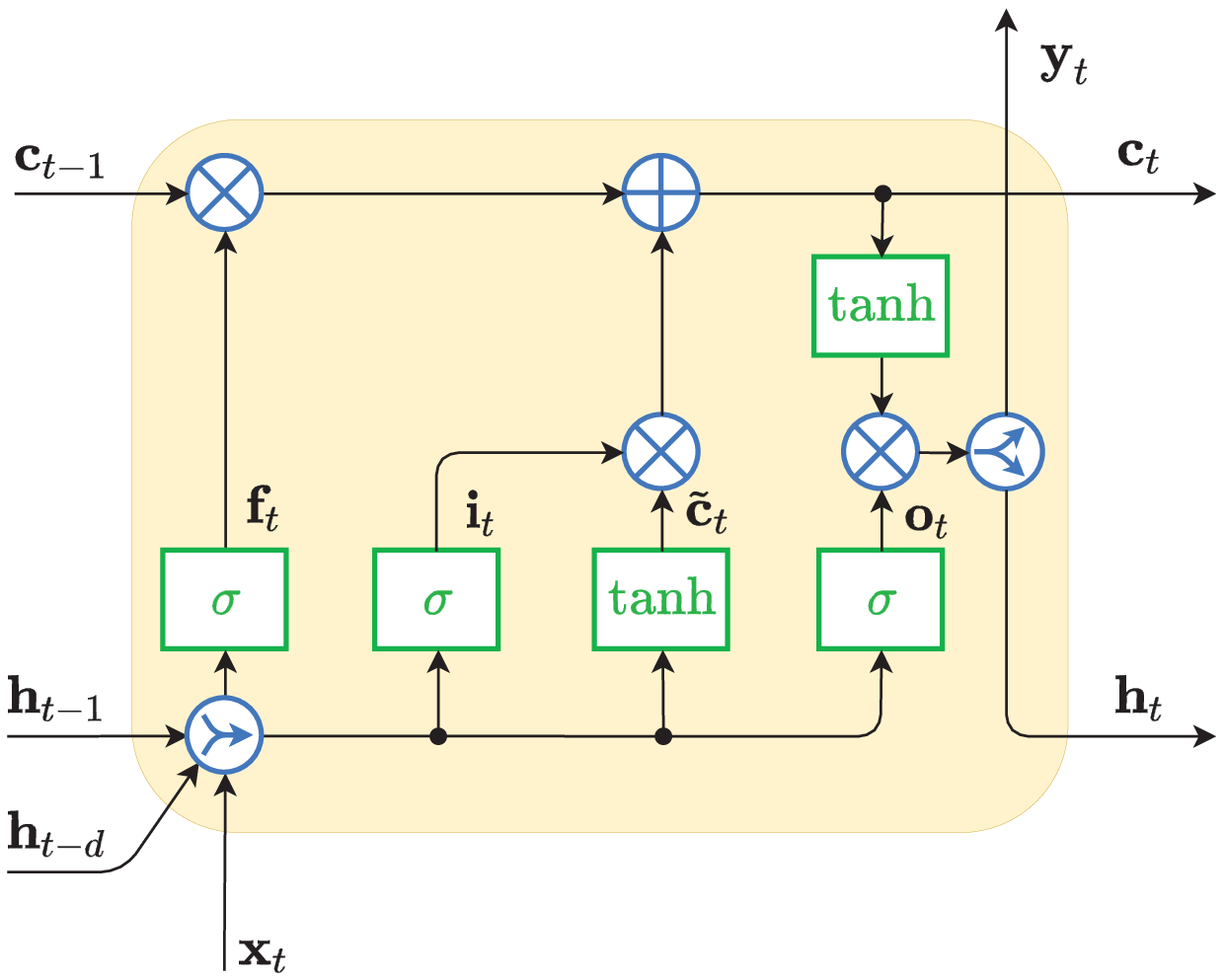}
        \hspace*{0.0cm}
        \includegraphics[align=c,width=0.50\textwidth]{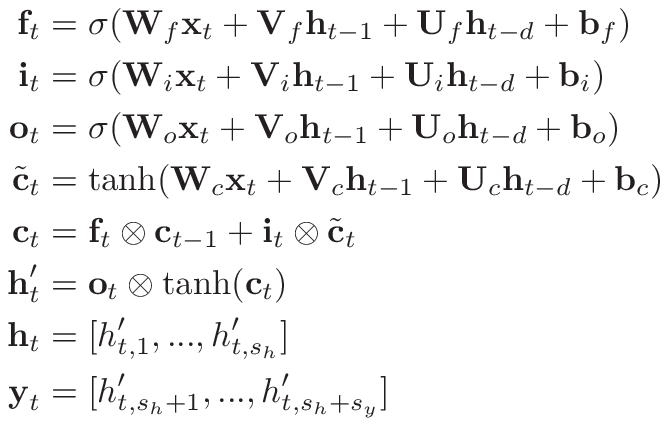}
    \end{minipage}
    \caption{dLSTM.}
    \label{figdLSTM}
\end{figure}

To improve the modeling of long-term dependencies in time series, we propose a dilated LSTM cell (Fig. \ref{figdLSTM}). Our modification comes down to two elements. First, in addition to the hidden state $\textbf{h}_{t-1}$, we introduce a delayed hidden state, $\textbf{h}_{t-d}$, $d>1$. This allows the data processing in time $t$ to be controlled using not only information from the recent state but also using direct information from the delayed state. This can be useful for seasonal time series, in which case the dilation can correspond to the period of seasonal variations.  
Second, the output hidden state, $\textbf{h}'_{t}$, is split into "real output" $\textbf{y}_t$, which goes to the next layer, and a controlling output $\textbf{h}_t$, which is an input to the gating mechanism in the following time steps. This solution was inspired by \cite{Ben17}. The size of the $c$-state is equal to the summed sizes of $h$-state and $y$-output, i.e. $s_c=s_h+s_y$. 

The equations corresponding to dLSTM are shown in Fig. \ref{figdLSTM}, where: $\textbf{W}$, $\textbf{V}$ and $\textbf{U}$ are learned weight matrices, $\textbf{b}$ are learned bias vectors, and $s_h, s_y$ are the lengths of hidden state and output vectors, respectively. 

\subsection{dRNNCell}

\begin{figure}[h]
    \begin{minipage}{1\textwidth}
        \centering
        \includegraphics[align=c,width=0.40\textwidth]{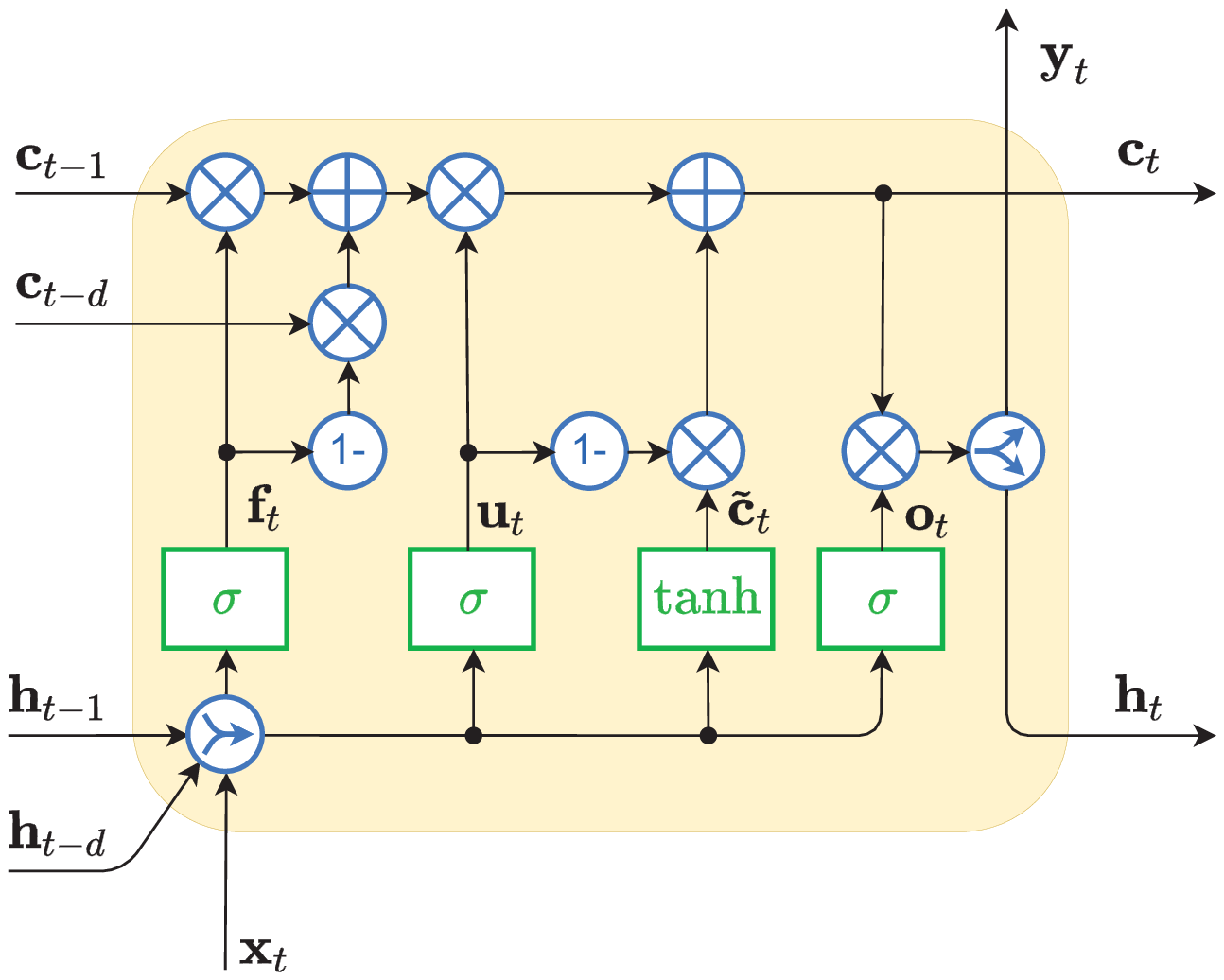}
        \hspace*{0.0cm}
        \includegraphics[align=c,width=0.50\textwidth]{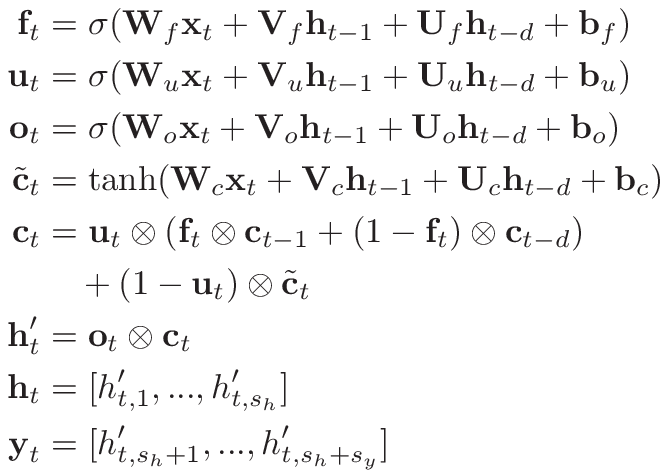}
    \end{minipage}
    \caption{dRNNCell.} 
    \label{figS2}
\end{figure}

A dilated recurrent NN cell, dRNNCell, was introduced in \cite{Smy21} as a combination of GRU and LSTM cells, see Fig. \ref{figS2}. It was designed to operate as part of a multilayer dilated RNN \cite{Cha17}. Its output is split into  $\textbf{y}_t$ and $\textbf{h}_t$ as in dLSTM. 

As in LSTM, dRNNCell uses two states, i.e. $c$-state and $h$-state. But, unlike LSTM, dRNNCell is fed by both most recent states, $\textbf{c}_{t-1}$ and $\textbf{h}_{t-1}$, and delayed states, $\textbf{c}_{t-d}$ and $\textbf{h}_{t-d}$, $d>1$. dRNNCell is equipped with three gates, which 
transform nonlinearly input vectors using logistic sigmoid function. They comprise fusion ($f$), update ($u$), and output ($o$) gates. A candidate $c$-state, $\tilde{\textbf{c}}_t$, is produced by transforming input vectors using $\tanh$ nonlinearity. The operation of the cell is described by the equations shown in Fig. \ref{figS2}. 

Note that the $c$-state is a weighted combination of past $c$-states and new candidate state $\tilde{\textbf{c}}_t$ computed in the current step. Update vector, $\textbf{u}_t$, decides in what proportion the old and new information are mixed in the $c$-state, while fusion vector $\textbf{f}_t$ decides about the contribution of recent and delayed $c$-states in the new state.

\subsection{adRNNCell}

An attentive dilated recurrent NN cell, adRNNCell, was proposed in \cite{Smy22} as an extended version of dRNNCell. It combines two dRNNCells to obtain a more efficient cell, which is able to preprocess dynamically the sequence data. It is equipped with an attention mechanism for weighting the input information. 

\begin{figure}[h]
    \begin{minipage}{1\textwidth}
        \centering
        \includegraphics[align=c,width=0.40\textwidth]{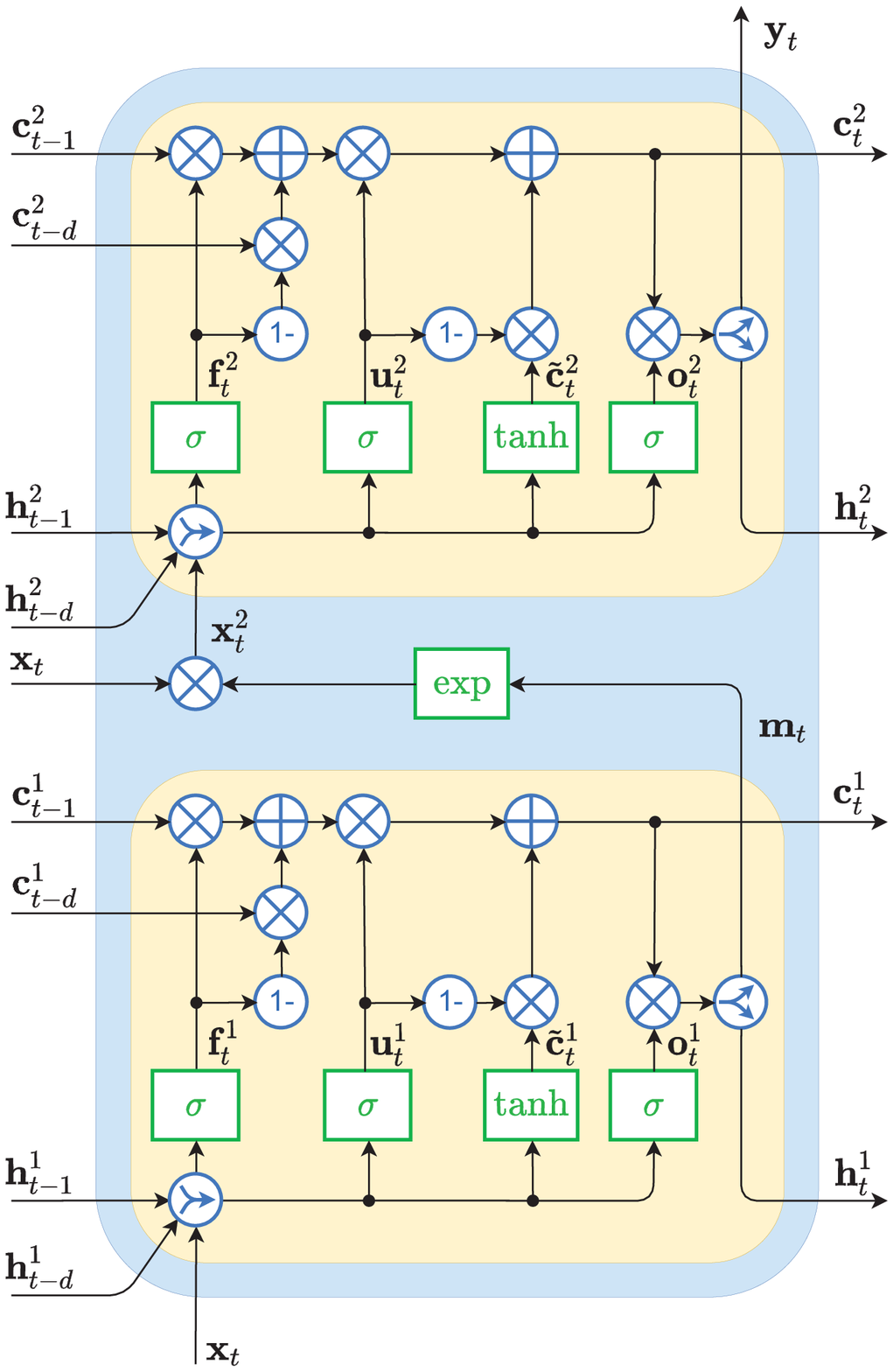}
        \hspace*{0.0cm}
        \includegraphics[align=c,width=0.50\textwidth]{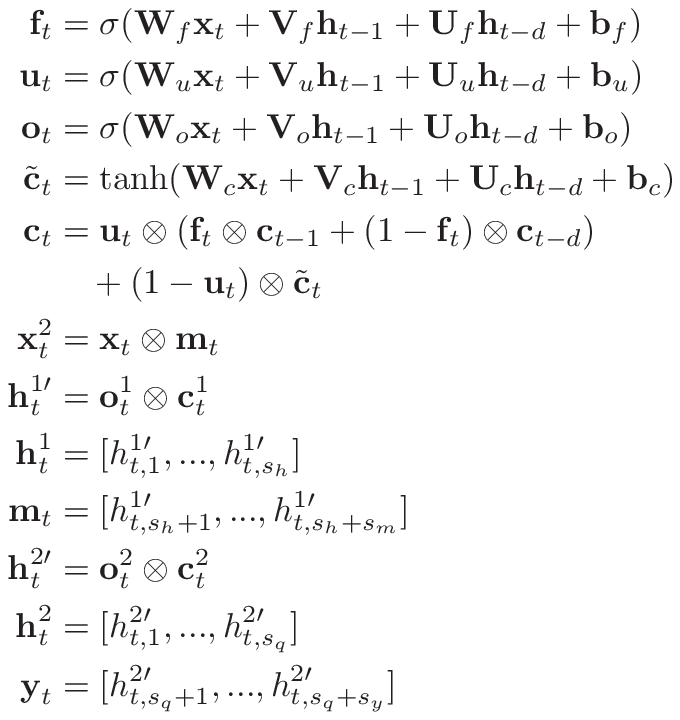}
    \end{minipage}
    \caption{adRNNCell.} 
    \label{figS3}
\end{figure}

Fig. \ref{figS3} shows adRNNCell composed of lower and upper dRNNCells. The former produces attention vector $\textbf{m}_t$ of the same length as the input vector $\textbf{x}_t$. The components of $\textbf{m}_t$, after processing by $\exp$ function, are treated as weights for the inputs collected in $\textbf{x}_t$. The weighted inputs, $\textbf{x}^2_t$, feed the upper cell. The goal of such an attention mechanism is to dynamically strengthen or weaken particular inputs depending on their relevance. Note that this process is dynamic, the weights are adjusted to the current inputs at time $t$.      
Both cells, lower and upper, learn simultaneously. Based on the weighted input vector, $\textbf{x}_t^2$, the upper cell predicts vector $\textbf{y}_t$.

The mathematical model describing adRNNCell is shown in Fig. \ref{figS3}.

\section{RNN Architecture}

\begin{figure}[h]
	\centering
    \includegraphics[width=0.6\textwidth]{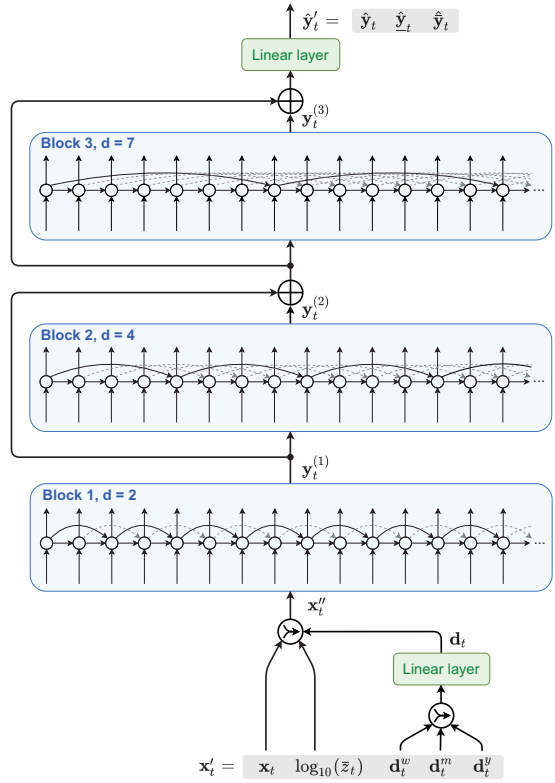}
    \caption{RNN architecture.} 
    \label{figRNN}
\end{figure}

In this study, we adopt RNN architecture from \cite{Smy22}. It is composed of three single-layer blocks, see Fig. \ref{figRNN}. In each block, the cells are dilated differently, i.e. 2, 4 and 7, respectively.
Delayed connections enable the direct input into the cell of information from a few time steps ago. This can be useful in modeling seasonal dependencies. 
To model the temporal dependencies of different scales, our architecture has multiple dilated recurrent layers stacked with hierarchical dilations. It also uses ResNet-style shortcuts between blocks to improve the learning process \cite{He16}.

To reduce input dimmensionality, the calendar variables, $\textbf{d}_t^{w}$, $\textbf{d}_t^{m}$ and $\textbf{d}_t^{y}$, are embedded using a linear layer into $d$-dimensional continuous vector $\textbf{d}_t$. The second linear layer at the top of stacked recurrent layers, produces the point forecasts, $\hat{\textbf{y}}_t$, and two vectors of quantiles, a lower one, $\hat{\underline{\textbf{y}}}_t \in \mathbb{R}^{24}$, and an upper one, $\hat{\bar{\textbf{y}}}_t \in \mathbb{R}^{24}$. These quantiles of assumed orders, $\underline{q}$ and $\overline{q}$, define the PI.

To enable our RNN to learn both point forecasts and PI quantiles, we employ the following loss function \cite{Smy21}: 

\begin{equation}
L =
\rho(y, \hat{y}_{q^*}) + \gamma(
\rho (y, \hat{y}_{\underline{q}}) + 
\rho (y, \hat{y}_{\overline{q}}))
\label{eqlss}
\end{equation}
where $\rho$ is a pinball loss: 

\begin{equation}
\rho(y, \hat{y}_q) =
\begin{cases}
(y-\hat{y}_q)q       & \text{if } y \geq \hat{y}_q\\
(y-\hat{y}_q)(q-1)  &\text{if } y < \hat{y}_q 
\end{cases}
\label{eqrho}
\end{equation}
$q \in (0, 1)$ is a quantile order, $y$ is an actual value (standardized), $\hat{y}_q$ is a forecasted value of $q$-th quantile of $y$, $q^*=0.5$ corresponds to the median, $\underline{q} \in (0,q^*)$ and $\overline{q} \in (q^*, 1)$ correspond to the lower and upper bound of PI, respectively, and $\gamma \geq 0$ is a parameter controlling the impact of the components related to PI on the loss function, typically between 0.1 and 0.5.

The first component in \eqref{eqlss} is a symmetrical loss for the point forecast, while the second and third components are asymmetrical losses for the quantiles. The asymmetry level, which determines PI, results from the quantile orders. For example, we obtain a 90\% symmetrical PI for $\underline{q} = 0.05$ and $\overline{q} =0.95$.  

Remarks:
\begin{enumerate}
\item
In Section \ref{ES}, we compare RNN with the different cell types which were described in Section \ref{RC}. Note that RNN shown in Fig. \ref{figRNN} requires cells equipped with both recent and delayed connections. Classical LSTM and GRU are not equipped with delayed inputs. RNN with these cells are considered in two variants: (i) LSTM1, GRU1 -- the delayed connections are removed and cells are fed with only recent inputs $t-1$, and (ii) LSTM2, GRU2 -- the recent connections are removed and cells are fed with only delayed inputs, $t-2$, $t-4$ or $t-7$, depending in the layer.

    \item
The pinball loss gives the opportunity to reduce the forecast bias by penalizing positive and negative deviations differently. When the model tends to have a positive or negative bias, we can reduce the bias by introducing $q^*$ smaller or larger than $0.5$, respectively (see \cite{Smy20,Dud21}).

\end{enumerate} 

\section{Experimental Study}
\label{ES}

We compare the performance of the proposed RNN with different recurrent cells on STLF problems for 35 European countries. The data, collected from ENTSO-E repository (www.entsoe.eu/data/power-stats), concerns real-world hourly electrical load time series.
The data period is from 2006 to 2018 but a large amount of data is missing in this period (about 60\% of the countries have complete data). For 35 countries, the data provides a variety of time series with triple seasonality expressing different properties such as different levels, trends, variance and daily shapes (see Section II in \cite{Smy21} where these time series are analysed). We treat data from 2018 as test data. We predict daily load profiles for each day of the test period and each country with the exception of three countries. For these three countries, due to missing data, the test periods were shorter, i.e. for Estonia and Italy (missing last month of data) and Latvia (missing last two months of data).

The RNNs were optimized on data from the period 2006-17. As performance metrics we use: mean absolute percentage error (MAPE), median of APE (MdAPE), interquartile range of APE (IqrAPE), root mean square error (RMSE), mean PE (MPE), and standard deviation of PE (StdPE). Below, we report results for an ensemble of five RNNs (average of five RNN runs).
We use a similar training and optimization setup as in \cite{Smy22}. The key hyperparameters were: $s_c=250$, $s_h=s_q=s_y=125$, $q^*=0.5$, $\underline{q} =0.05$, $\overline{q}=0.95$, $\gamma=0.3$, number of epochs: 10, learning rates: $3\cdot10^{-3}$ (epochs 1-5), $10^{-3}$ (epoch 6), $3\cdot10^{-4}$ (epoch 7), $10^{-4}$ (epochs 8-10), batch size: 2 (epochs 1-3), 5 (epochs 4-10).  

Table \ref{tabEr} displays the forecasting quality metrics averaged over the 35 countries. The results indicate that, on average, adRNNCell is the best cell according to three accuracy measures, MAPE, MdAPE and RMSE. It also produces the least dispersed forecasts -- see the lowest values of IqrAPE and StdPE. The second most accurate and precise cell is dRNNCell. The worst results are for GRU1.

\begin{table}[h]
	\setlength{\tabcolsep}{9pt}
	\caption{Forecasting quality metrics.}
	\begin{center}
	\begin{tabular}{lcccrrc}
		\toprule
		Cell type& MAPE  & MdAPE & IqrAPE & RMSE  & MPE   & StdPE \\
		\midrule    
    GRU1 & 2.31  & 2.10  & 2.23  & 318.69 & \textbf{-0.06} & 3.86 \\
    GRU2 & 2.26  & 2.04  & 2.19  & 308.92 & -0.15 & 3.78 \\
    LSTM1 & 2.25  & 2.03  & 2.18  & 307.09 & -0.19 & 3.78 \\
    LSTM2 & 2.16  & 1.94  & 2.10  & 293.00 & -0.10 & 3.60 \\
    dLSTM & 2.19  & 1.97  & 2.12  & 297.58 & -0.19 & 3.66 \\
    dRNNCell & 2.15  & 1.93  & 2.09  & 292.60 & -0.15 & 3.57 \\
    adRNNCell & \textbf{2.12} & \textbf{1.91} & \textbf{2.07} & \textbf{289.32} & -0.14 & \textbf{3.52} \\
		\bottomrule
	\end{tabular}
	\label{tabEr}
	\end{center}
\end{table}

\begin{figure}[h]
	\centering
	\includegraphics[width=0.4\textwidth]{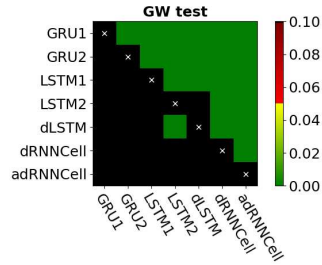}
	\caption{Results of the Giacomini-White test.} 
	\label{GW}
\end{figure}

To confirm the performance of adRNNCell, we perform a pairwise one-sided Giacomini-White test (GM test) for conditional predictive ability \cite{Gia06} (we used the multivariate variant of the GW test implemented in https://github.com/\\
jeslago/epftoolbox \cite{Lag21}). Fig. \ref{GW} shows the obtained $p$-values of this test. The closer the $p$-values are to zero
the significantly more accurate the forecasts produced by the model on the $X$-axis are than the forecasts produced by the model on the $Y$-axis. The black color is for $p$-values larger than 0.10, indicating rejection of the hypothesis that the model on the $X$-axis is more accurate than the model on the $Y$-axis. Fig. \ref{GW} clearly shows that adRNNCell and dRNNCell performed best. 

\begin{figure}[h]
	\centering
	\includegraphics[width=0.458\textwidth]{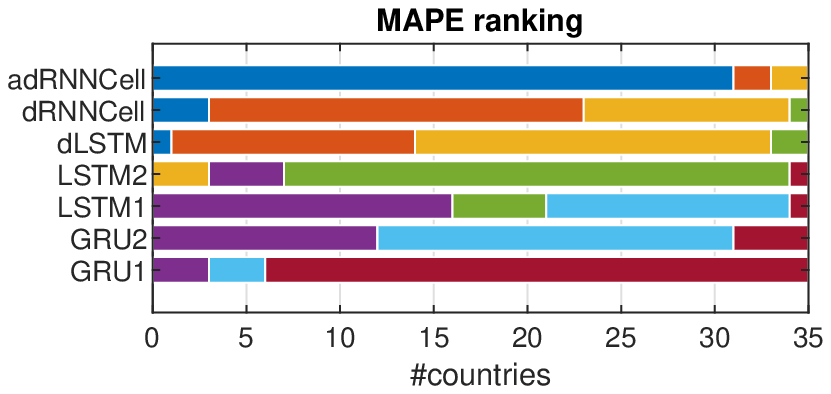}
	\includegraphics[width=0.53\textwidth]{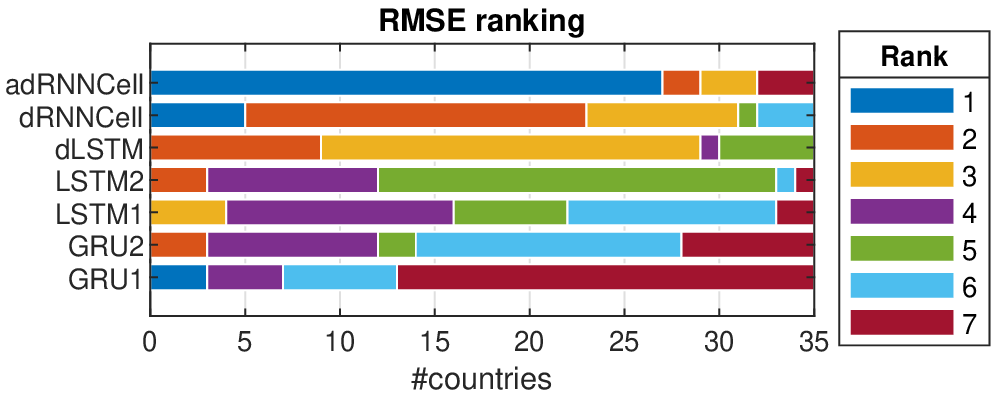}
	\caption{Results of MAPE and RMSE rankings.} 
	\label{Rank}
\end{figure}

Fig. \ref{Rank} shows rankings of the examined RNNs based on average errors for each country. Note the high position of adRNNCell. For 31 out of 35 countries this model gave the lowest MAPE, and for 27 countries it also gave the lowest RMSE. The second highest ranked model was dRNNCell.   

In Table \ref{tabEr}, we also show MPE, which is a measure of the forecast bias. Its negative values for all cases indicate over-prediction. The proposed model, thanks to the pinball-type loss function, can control the bias. For example, to reduce the bias for dRNNCell and adRNNCell we assumed $q^*=0.485$. This resulted in a reduction of MPE to $-0.04$ without decreasing the forecast accuracy. So, it is possible to reduce the biases shown in Table 1 further, but we were trying to prevent over-tuning of the hyperparameters, so left them as they are reported.

\begin{figure}[h]
	\centering
	\includegraphics[width=1\textwidth]{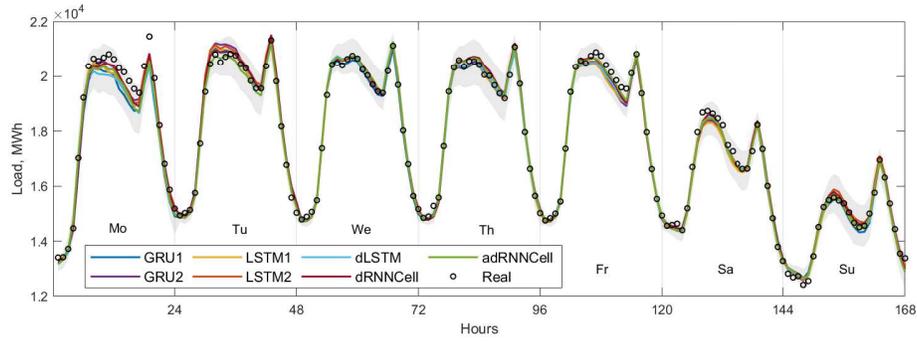}
	\caption{Examples of the forecasts. 90\% PIs for adRNNCell are shown as gray-shaded areas.} 
	\label{figff}
\end{figure}

Fig. \ref{figff} shows example forecasts of daily profiles for different days of the week. Note that forecasts generated by RNN with different cells do not differ much from each other. Fig. \ref{figff} also shows PIs for adRNNCell. To evaluate the accuracy of the PIs, we calculated the percentage of forecasts lying inside, above and below their PIs. The results are shown in Table 2. The predicted 90\% PIs cover the forecasts most accurately in the case of GRU2. But note that our loss function \eqref{eqlss} gives us the opportunity to tune further PIs. This can be performed by adjusting the quantiles determining the PI bounds, $\underline{q}$ and $\overline{q}$. 

In Table 2, a Winkler score is also shown. For observations that fall within the PI, this score is simply the length of the PI, while for observations outside PI, the penalty applies, which is proportional to how far the observation is outside PI \cite{Hyn22}. 
To bring the Winkler scores for different countries to a comparable level, we divide these scores by the mean loads of the corresponding countries in the test period. Such unified Winkler scores are shown in Table 2. Note that adRNNCell has the lowest Winkler score and adRNNCell the second lowest.

\begin{table}[h]
    \setlength{\tabcolsep}{9pt}
    \caption{Forecasting quality metrics.}
    \begin{center}
    \begin{tabular}{lcccc}
        \toprule
        Cell type  & \% in PI & \% below PI & \% above PI & Winkler score \\
        \midrule   
        GRU1 & 92.78±2.80 & 3.19±1.33 & 4.03±1.68 & 0.1524±0.2579 \\
    GRU2 & \textbf{90.40±3.40} & \textbf{4.79±1.74} & \textbf{4.81±1.80} & 0.1448±0.2710 \\
    LSTM1 & 89.16±4.07 & 5.30±2.05 & 5.53±2.21 & 0.1428±0.2760 \\
    LSTM2 & 88.52±3.88 & 5.53±1.96 & 5.96±2.04 & 0.1368±0.2741 \\
    dLSTM & 88.84±3.74 & 5.59±1.96 & 5.57±1.94 & 0.1393±0.2737 \\
    dRNNCell & 87.51±3.54 & 6.16±1.92 & 6.33±1.77 & 0.1363±0.2771 \\
    adRNNCell & 88.41±3.14 & 5.68±1.67 & 5.91±1.62 & \textbf{0.1332±0.2683} \\
 
        \bottomrule
    \end{tabular}
    \label{tabErr}
    \end{center}
\end{table}

Our research shows that adRNNCell is the best gated cell for forecasting time series with multiple seasonality. In \cite{Smy22} we compared RNN based on adRNNCells with a variety of forecasting models including statistical models (ARIMA, exponential smoothing, Prophet) and ML models  (MLP, SVM, ANFIS, LSTM, GRNN, nonparametric models). This comparison clearly showed that the adRNNCell-based approach outperforms all its competitors in terms of accuracy.     

\section{Conclusion}

In this study, we explore the potential of RNNs with different cells for forecasting time series with multiple seasonality. The best RNN solutions use dRNNCells and adRNNCells, cells designed especially for such complex time series. They outperform classical GRU and LSTM cells as well as modified LSTM with dilation. adRNNCell, which is the most advanced cell with dilation and attention, combines two dRNNCells: one of which learns an attention vector while the other uses this vector to weight the inputs. The attention mechanism enables the cell to preprocess dynamically the sequence data while the delayed connections enable it to capture the long-term and seasonal dependencies in time series. 

Apart from the dilation and attention mechanisms, the superior performance of the proposed RNN has its sources in the following mechanisms and procedures. First is the multilayer architecture, which is composed of several dilated recurrent layers stacked with hierarchical dilations to deal with multiple seasonality. Second is cross-learning on many time series, which enables RNN to capture the shared features of the individual series and helps to avoid over-fitting. 
Third is a time series representation using standardized weekly patterns as inputs and encoded daily patterns as outputs. The encoding variables are determined from the history, which enables decoding. Fourth is a composed asymmetrical loss function based on quantiles, which enables RNN to produce both point forecasts and PI and also to reduce the forecast bias.

In further research, we plan to enrich the input information with a learned context vector. This represents information extracted from other time series, which can help predict a given time series.


\end{document}